# Spatiotemporal Ground Reaction Force Analysis using Convolutional Neural Networks to Analyze Parkinsonian Gait

Musthaq Ahamed[1], P.D.S.H. Gunawardane[2], Nimali T. Medagedara[1]

[1] The Open University of Sri Lanka, Nawala, Nugegoda.
[2] The University of British Columnbia, Vancouver, Canada.
musthaqahamed02@gmail.com, hiroshan@mail.ubc.ca, tmmed@ou.ac.lk

**Abstract-** Parkinson's disease (PD) is a non-curable disease that commonly found among elders that greatly reduce their quality of life. PD primarily affects the gait pattern and slowly changes the walking gait from the normality to disability. The early diagnosing of PD is important for treatments and gait pattern analysis is used as a technique to diagnose PD. The present paper has identified the raw spatiotemporal ground reaction force (GRF) as a key parameter to identify the changes in human gait patterns associated with PD. The changes in GRF are identified using a convolutional neural network through pre-processing, conversion, recognition, and performance evaluation. The proposed algorithm is capable of identifying the severity of the PD and distinguishing the parkinsonian gait from the healthy gait. The technique has shown a 97% of accuracy in automatic decision-making process.

## 1 Introduction

Parkinson's disease (PD) is a progressive neurodegenerative disorder that affects elderly population. The PD extirpate a small region in the brain that controls movement, balance, and posture. In 2016, 6.1 million people are diagnosed with PD [1]. The reports have shown a significant increase in reported Parkinson cases due to the aging of the world population. In first five to ten years, most of the Parkinson subjects have not shown any significant symptoms to early diagnose this disease [2].

The available techniques to diagnose PD are relying on the experience of human technicians [3]. Unified Parkinson Rating Scale (UPDRS) is the most common technique that is used to diagnose PD [4]. The UPDRS consists of five categories to diagnose the disease severity stages including behavior, physical movement (motor) examination, mood, mentation, and day to day activities. In one of the studies, PD is evaluated by interviewing the subjects [5]. However, the analysis of gait characteristics/patterns for PD is found by Knutsson in 1972 [6]. This study has shown that the parkinsonian gait has a large variability compared with healthy gait.

Moreover, PD subjects have lower walking speed with a large gait cycle. The parkinsonian gait's stride length is shorter compare to the healthy gait. This specificity in the gait pattern is caused by the freezing of the gait and it is globally identified as a vital feature of identifying the PD [7]. Even though human gait movement and their posture are unique, in some cases these technicians have made mistakes in distinguishing parkinsonian gait [8]. Therefore, gait analysis of subjects with PD requires a highly sophisticated approach to get reliable results. Deep learning approaches are used to enhance the reliability in gait analysis [9].

In recent years, artificial neural networks and deep learning has been advanced to process and analyze complex data sets [10], [11]. The development of computer technologies has introduced new approaches for data processing, analyzing, and data classifying in healthcare applications [12]. The deep learning is used to automate the development of models that is used to analyze complex ground reaction force (GRF) data to deliver faster and detailed results in gait related applications [13]. These approaches are modifiable to effectively identify the gait abnormalities using spatiotemporal gait parameters. The present paper explores the performance of the deep neural network to classify parkinsonian and normal gait patterns.

## 1.1 Literature review

The sensitivity of the GRF data is crucial to classification of parkinsonian gait from normal gait. Pressure sensing is used to measure GRF generated in the foot during walking. A similar technique has been used to classify gait patterns among the children with Autism using ANN and support vector machine [14]. This paper has reported of technique to classify Autism gaits with a 95% of accuracy. This result has validated the effective use of machine learning approaches to classify pathological gait patterns. In [15], deep learning has been used with GRF data for wide area floor sensing. This study has explored the techniques of categorizing gait patterns by fusing raw spatiotemporal GRF. The paper has explored the use of CNN with long short-term memory approaches to study parkinsonian gaits. This technique has shown a 96% precision of identifying these gait patterns.

The technique introduced in [16] has used a computer vision-based technique to identify parkinsonian gait and a 95.49% accuracy has been reported. This study has been compared the normal gaits with parkinsonian gaits and has been able to successfully identify several progressive stages. Therefore, one of the primary goals of the present paper is to classify the parkinsonian gait based on the disease severity. Several previous attempts have used machine learning, vision-based systems, deep learning models, and statistical methods to classify parkinsonian gaits [14], [15], [16], [17], [18]. However, the classification of severity stages of parkinsonian gait is highly important for the treatments and have not been covered in many of these studies. Therefore, this paper has focused on developing a cost effective and fast data processing application based on CNN to detect parkinsonian gaits to be apply for future medical applications.

This paper is structured as follows. Section 2 summarizes the database that is used for this study. Section 3 explains the methodology; data preprocessing, data matrix form development, image conversion, data preparation for CNN model, pattern recognition model development, and performance evaluation. Section 4 presents the

CNN model performance and discusses the results and section 5 concludes the final outcome and presents recommendations for the future work.

## 2  Data Description

The data set used in this study was obtained from the open-source database PhysioNet [19]. The data included 93 idiopathic PD subjects and 73 healthy subjects with an average age of 66. This data consists PD disease severity measurement Hoehn and Yahr score for each subject. There were 10 subjects with the Hoehn and Yahr PD stage 3, 27 subjects with a Hoehn and Yahr PD stage 2.5, and 56 subjects with a Hoehn and Yahr PD stage 2. The dataset contains vertical GRF for both PD and healthy subjects with approximately 2 minutes of walking. The GRF was gathered using three separate measures, which are known as dual tasking (Ga), rhythmic auditory stimulation (Ju), and treadmill walking (Si) which is shown in Table 1. The dual tasking is used to evaluate the limitations of the attentional capability of the subjects [20]. The rhythmic auditory stimulation improves the gait and gait-related movements [21]. Similarly, the treadmill walking improves the gait and gait-mobility [22].

**Table 1.** Data description of the number of participants.

| Group | Subjects | |
|---|---|---|
| | *Number of PD subjects* | *Number of healthy subjects* |
| Ga | 29 | 18 |
| Ju | 29 | 26 |
| Si | 35 | 29 |

The data acquisition for every subject is carried out using eight flex sensors under each foot to capture the force applied to it during walking. Each sensor produced 100 samples per second and the readings are recorded in a raw matrix. After the data acquisition, all eight-sensor data from each foot is added and recorded separately into the matrix with their respective subjects [19].

## 3  Methodology

This section describes the data preprocessing, image conversion, development of the CNN and performance evaluation techniques.

### 3.1  Data preprocessing

The raw data files were converted from text to CSV for each sample and saved separately prior to the preprocessing. Each sample had 19 columns and data were recorded in a slightly different frame rate (some data had 12119 and others had 1000 frames per sample). The columns included were; time stamps, GRF data from the left foot of the each subject (eight sensor measurements), GRF data form the right foot of the each subject (eight sensor measurements), and the two total GRF values for each foot. The frame rate mismatch was handled as in [23], [24]. The final preprocessed data set after omitting the unnecessary data for CNN model, each subject data contained 18 columns and 500 frames. A single gait cycle including both heel strike and toe-off takes one second (100 frames), therefore, each subject data contained 5 gait cycles (i.e., GRF of each subject for 5 gait cycles). These samples were converted into a 500x18 matrices. The number of samples for each data class is as shown in the Table 2. The classified data for PD stage 2 and PD stage 3 had the maximum and the minimum number of samples respectively [25].

**Table 2.** Number of samples used for each classification

| Data class | Number of samples |
|---|---|
| Healthy subjects | 2001 |
| PD stage 2 subjects | 2084 |
| PD stage 2.5 subjects | 1633 |
| PD stage 3 subjects | 541 |

### 3.1 Data Normalization

The data normalization is critical to deep learning models to map their inputs with the outputs. The differences in the scales must be omitted and data should be comparable with each other to corelate and identify their characteristics. This data set is normalized between 0 to 1 [26]. Each 500×18 matrix has considered independently in the data normalization process. After the normalization, the matrices are represented as an image with their respective color shades. The created spectrogram is contributed each with a height of 500 pixels and a width 18 pixels, totaling 9000 features. The reconstructed image samples are shown in Fig. 1.

The spectrograms represent the GRF of the data set; healthy subjects, PD stage 2 subjects, PD stage 2.5 subjects, and PD stage 3 subjects respectively from left to right. Similarly, 6259 images are created to represent all the data. The last two sections (columns) of the image are the total GRF of left and right foot respectively. Comparing to the samples of the healthy subjects, the PD subjects have shown a high total GRF due to their freezing of the gait during the walking.

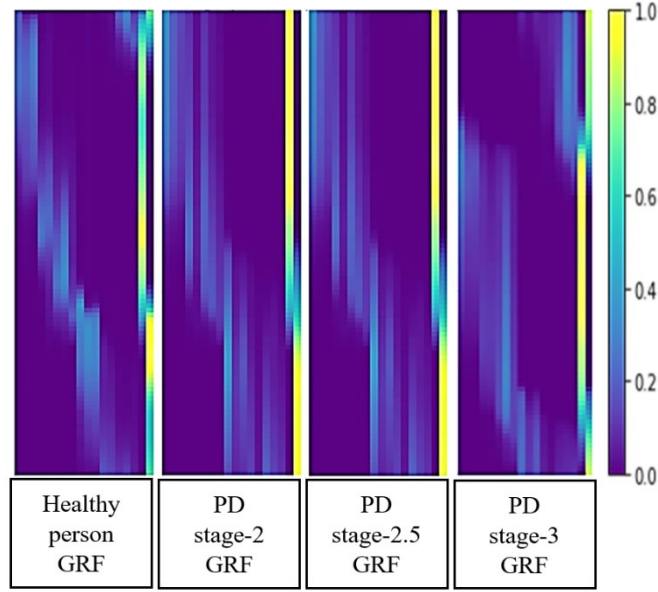

**Fig. 1.** The normalized image samples. The yellow and purple colors represent the maximum and minimum GRF

### 3.2 Pattern Recognition Model Development

The data set is unstructured and labelled, therefore, supervised learning is suitable to be used in this application. Supervisory learning maps the data sets with inputs and outputs. The primary goal of this algorithm is to seek the relation between input images and GRF group labels. However, there is no quantifiable parametric variation between the gait patterns of PD and healthy subjects. Therefore, to identify the relation between GRF and PD, this proposed technique in this paper is used. Moreover, the developed CNN model can identify the PD stages concerning their Hoehn Yahr PD progress scales. The CNN model is used to learn the events and the phase changes during the gait cycles [27].

### 3.3 Convolution Neural Network (CNN)

CNN's capability of automatically grasping sophisticated patterns from large datasets is used for this application. In general, CNN consists of a convolution layer and a pooling layer [28], [29]. The images are initially fed into convolution layer which performs the convolution operation. The input convolution layer is set to 500×18×1 to fit with image size, including a single-color channel. This is a linear operation that multiplies set of weights with the input image's pixel values. The multiplication is achieved between an input pixel data array and two-dimensional weight array that is known as a filter [30]. The filters are used to determine the features that associate with each image. The convolution operation used in this algorithm is for a two- dimensional image and shown in the Equation 1 [31].

$$S(i,j) = (I \times K)(i,j) = \sum_m \sum_n I(m,n)K(i-m, j-n) \qquad (1)$$

The parameters and variables are; two-dimensional image $= I$, two-dimensional filter $= K$, the indexes of rows and columns of the result matrix are marked with $i$ and $j$ respectively.

After applying the filter over a selected set of pixels, each value of the filter is multiplied with their corresponding value from the input image. Then all the multiplied pixels are summed up and placed in the output feature map. The features of the images such that edges and corners are identified using this convolution layer. Specifically, the generated output feature maps are maintained same size of the input data matrix by adding a padding value as a one into the convolution operation. And also, the rectifier linear function is applied to increase the non-linearity in feature map because the images have nonlinear features such as tractions between pixels, boarders, and colors. The model is constructed using four convolution layers with three hidden layers. The filter selection for each layer is given in Table 3. The filters are chosen concerning the maximum accuracy with the best performance characteristics [32].

**Table 3.** Number of filters used for each layer

| Layer description | Number of filters |
| --- | --- |
| Convolution layer | 128 |
| 1st hidden convolution layer | 256 |
| 2nd hidden convolution layer | 512 |
| 3rd hidden convolution layer | 1024 |

The filters are selected in incremental order with the size of 3×3 to keep the feature space wide and shallow from the initial stages of the network while making it narrow and deep towards the end of the model. Every convolution layer is followed by pooling layer that reduces the spatial size of the representation [33]. As a result, the number of parameters and computation time in the network are minimized.

The pooling operation is carried out by sliding a two-dimensional filter over each channel of the feature map and then the features are summarized within the region that covered by the filter. After, pooling operation, the dimensions of the feature map $n_h \times n_w \times n_c$ are obtained using the Equation 2 [34].

$$y = \left(\frac{n_h + 2p - f}{s} + 1\right) \times \left(\frac{n_w + 2p - f}{s} + 1\right) \times n_c \qquad (2)$$

Where, $y$ is the dimension of output after the pooling operation, $n_h$ is the height of feature map, $n_w$ is the width of the feature map, $n_c$ is the number of channels in the feature map, $f$ is the size of the filter, $s$ is the stride length, and $p$ is the padding.

In this model, max pooling is selected by the pooling operation that selects the maximum element from the region of the feature map covered by the filter with the size of a 2×2 matrices [35]. The padding is selected to be zero by less considering the edges of the images. Similarly, to reduce the timed consumed for operation and complexity, the stride length is selected as 2. The convolution matrix of $i^{th}$ output after the max-pooling is a feature map containing the most outstanding features of the previous feature map. This is shown in Table 4. For instance, after the 1$^{st}$ pooling operation, the generated output feature map size is determined using Equation 2. The height and width of feature map after the 1$^{st}$ convolution operation is 500×18. The other parameters are selected to be $p$=0, $s$=2, $f$=2, and $n_c$=1 (single color channel). By substituting the parameters into Equation 2, the generated output shape turned out to be 250×9×1. Similarly, each output feature map is produced using the same technique.

**Table 4.** The output image shape for each pooling operation.

| Layer description | Output shape |
|---|---|
| 1$^{st}$ pooling layer | 250 × 9 |
| 2$^{nd}$ pooling layer | 125 × 4 |
| 3$^{rd}$ pooling layer | 62 × 2 |
| 4$^{th}$ pooling layer | 31 × 1 |

In general, overfitting provides too much information to CNN, specifically, by the information that is irrelevant to the model. Therefore, overfitting can be avoided by constructing a pooled feature map in the model [36]. After the pooled featured map is obtained, the fatten function is used to transform the entire pooled feature map matrix into a single column that could be fed into the neural network for classification. Flattening has resulted a single long feature vector that is used for the neural network. After flattening, the flattened feature map has been passed through another neural network. This step consists of an input layer, a fully connected layer, and an output layer. The fully connected layer is same as the hidden layer in ANNs. However, in this algorithm, it's fully connected. The output layer is the predicted class of each gait pattern [37], [48].

The identified pooled features are passed through the network and the error of prediction is calculated. The fully connected layer is developed using 512 neurons that consist of rectified linear unit (ReLU) activation function. Similarly, the output layer contains 4 neurons with softmax activation function that represents each output of the PD case's gait pattern. The softmax activation function is selected for the output layer to represent the probability of each class that brought down to numbers between zero and one. The model loss is calculated using categorical cross-entropy that could handle multi-class classification [39], [40].

The developed CNN model is optimized using an adaptive learning rate (Adam) optimization algorithm that is used to obtain individual learning rates of each parameter. The 80-percent of the dataset is used to train the model and the remaining is used to test and validate the model. The callback, early stopping functions are used to stop training when it reaches a certain accuracy with loss score by adjusting the learning rates over time and to prevent the overfitting during the training stage. Fig. 2 shows the proposed 2D-CNN architecture with the descriptions of each layer. The outcome of the model is determined by using human gait pattern analysis that specifies for each case [41]. CNN model is developed using open-source libraries that are found in SciKit learn, TensorFlow, and Keras (TensorFlow backend) [42].

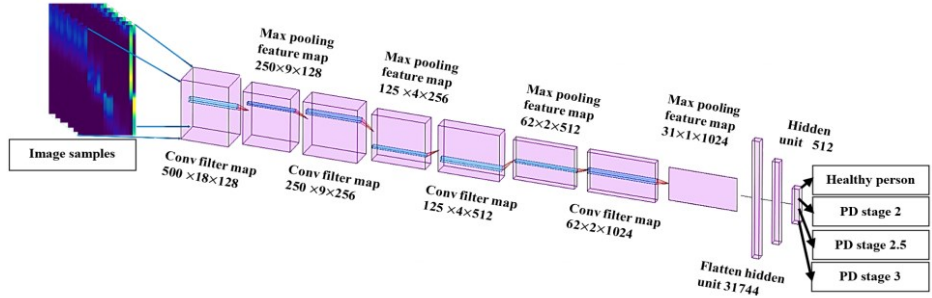

**Fig. 2.** Illustration of the proposed CNN architecture.

### 3.4 Model Performance Evaluation

The performance of the classification algorithm is determined by the confusion matrix shown in Fig. 3. The gait pattern recognition of each subject is analyzed through the confusion matrix [43], [44].

|   | Predicted class | |
|---|---|---|
|   | P | N |
| **Actual Class** P | True Positives (TP) | False Negatives (FN) |
| **Actual Class** N | False Positives (FP) | True Negatives (TN) |

**Fig. 3.** Confusion matrix.

Various performance measures are used to examine the model performance in the confusion matrix. Mainly, model accuracy, recall, precision, and F-measure parameters are obtained to evaluate the performance.

The model accuracy is determined by Equation 3. Where true positive = $TP$, true negative = $TN$, false positive = $FP$ and false negative = $FN$.

$$Accuracy = \frac{TP + TN}{TP + TN + FP + FN} \quad (3)$$

The number of true positive value is determined using the recall function shown in Equation 4.

$$Recall = \frac{TP}{TP + FN} \quad (4)$$

F-measure is used to measure the model performance concerning precision and recall. Equation 5 shows the F-measure calculation formula [48].

$$F-measure = \frac{2 \times Recall \times Precision}{Recall + Precision} \quad (5)$$

## 4 Results

The results showed the prediction capability of the model as mentioned in the previous section. Moreover, model training and testing characteristics were visualized using graphs. The progress of the training, testing accuracy, and losses were plotted concerning 12 epochs as shown in Fig. 4 and Fig. 5 respectively. The model confusion matrix shows the comparison between prediction class and actual class resulted as in Fig. 6. The overall model performance scores are summarized as shown in Table 5, including each PD stage analysis.

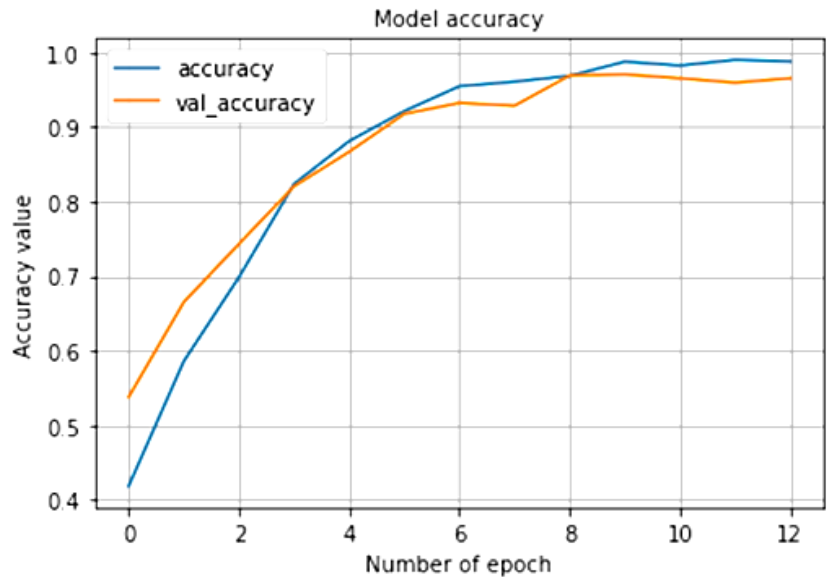

**Fig. 4.** Model training accuracy and validation accuracy plot.

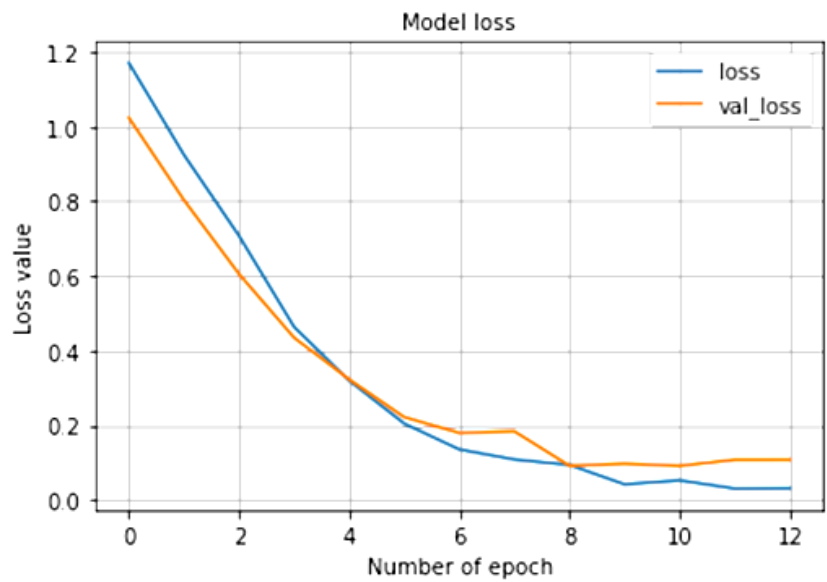

**Fig. 5.** Model training loss and validation loss.

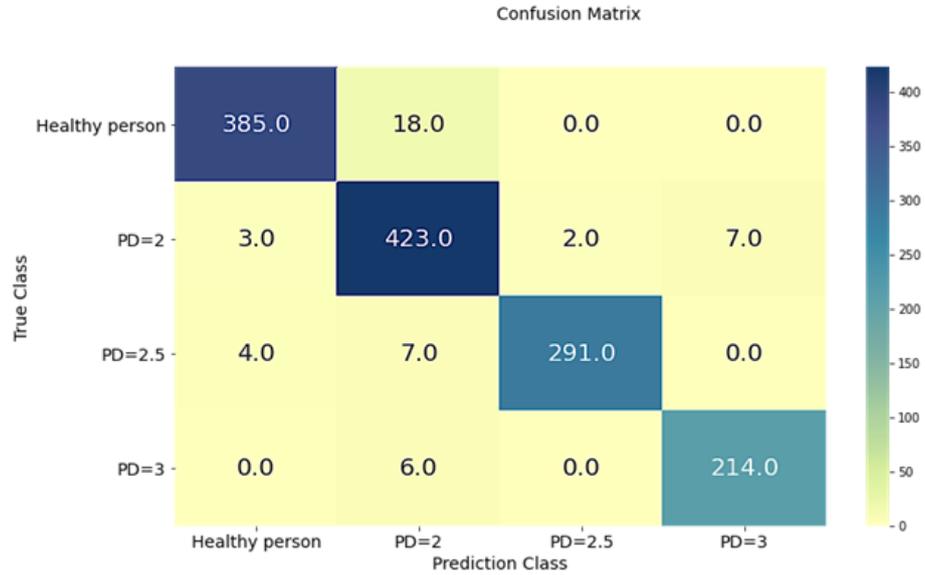

**Fig. 6.** CNN model confusion matrix.

**Table 5.** Model performance.

| Case | Precision (%) | Recall (%) | F1-measure (%) | Overall accuracy (%) |
|---|---|---|---|---|
| Healthy person | 98 | 96 | 97 | 97 |
| PD stage 2 | 93 | 97 | 95 | 97 |
| PD stage 2.5 | 99 | 96 | 98 | 97 |
| PD stage 3 | 97 | 97 | 97 | 97 |

The results have shown that the proposed methods are capable of accurately predict the parkinsonian gait and its stages. Table 5 shows CNN algorithm performed with a maximum peak accuracy value of 97%. This technique improved the accuracy by 1% in comparison to the technique introduced in [15]. Although when compares to [15] the accuracy had increased by a 1%. The parameters; precision, recall, and F1-measure values were recorded over 93% accuracy showing the model reliability. This work has provided evidence that having a higher number of filters can enhance the model performance characteristics in this application. Fig. 4 shows the progress of the training and validation accuracy has reached 97% and confirmed that accuracy of the model. Similarly, the model training and validation losses were analyzed as shown in Fig. 5, achieving less than 0.2 obtaining a good fit. Moreover, the PD stage 2 showed the lowest precision value by scoring 93%. A key characteristic shown in results is the higher value in the super diagonal matrix. This is an indication of high accuracy of prediction in the CNN model. The current systems require only few apparatuses with respect to the vision-bases system in [16].

## 5   Conclusion

GRF data is a crucial parameter in human gait analysis to identify PD. This paper had presented a CNN based approach to identify and classify parkinsonian gait and stages of PD using GRF-based data recorded in human trials. The proposed technique had shown a 97% accuracy of prediction and the lowest precision recorded for classifying different stages of PD was 93%. Therefore, this work suggests CNN as a best approach to identify PD and their stages. The future work of this research would be helped for investigate other gait parameters of parkinsonian gait (e.g., joint velocities/acceleration, walking speed, etc.) for early diagnosing of PD.